\documentclass[runningheads]{llncs}
\usepackage{graphicx}
\usepackage{amsmath,amssymb,graphicx,subfigure,multirow,float,makecell} 
\usepackage{booktabs}
\usepackage{color}
\usepackage[width=122mm,left=12mm,paperwidth=146mm,height=193mm,top=12mm,paperheight=217mm]{geometry}

\begin{document}
\pagestyle{headings}
\mainmatter

\title{Generative Adversarial Networks for Unsupervised Object Co-localization}
\titlerunning{Generative Adversarial Networks for Unsupervised Object Co-localization}

\authorrunning{Junsuk Choe, Joo Hyun Park, Hyunjung Shim}

\author{Junsuk Choe, Joo Hyun Park, Hyunjung Shim}
\institute{Yonsei University}

\maketitle

\begin{abstract}
This paper introduces a novel approach for unsupervised object co-localization using Generative Adversarial Networks (GANs). GAN is a powerful tool that can implicitly learn unknown data distributions in an unsupervised manner. From the observation that GAN discriminator is highly influenced by pixels where objects appear, we analyze the internal layers of discriminator and visualize the activated pixels. Our important finding is that high image diversity of GAN, which is a main goal in GAN research, is ironically disadvantageous for object localization, because such discriminators focus not only on the target object, but also on the various objects, such as background objects. Based on extensive evaluations and experimental studies, we show the image diversity and localization performance have a negative correlation. In addition, our approach achieves meaningful accuracy for unsupervised object co-localization using publicly available benchmark datasets, even comparable to state-of-the-art weakly-supervised approach.
\end{abstract}

\section{Introduction}
\label{sec:introduction}

Object localization aims to identify the location of a target object in a given scene. Recently, deep learning based methods, such as Faster R-CNN \cite{ren2017faster}, YOLO \cite{redmon2016yolo}, and SSD \cite{liu2016ssd}, have achieved significant improvement in object detection with real-time performance. These techniques \cite{ren2017faster,redmon2016yolo,liu2016ssd,krizhevsky2012imagenet,sermanet2013overfeat,simonyan2014very,szegedy2015going,he2016deep,girshick2015fast,girshick2014rich,redmon2016yolo9000}, however, utilize fully-supervised learning, which require category labels and bounding box annotations for training. Because such information is considered expensive, acquiring massive amount of data annotations is difficult, drawing a limit for practical applications.

To alleviate the burden on data annotations, weakly-supervised learning methods have been suggested. The weakly-supervised object localization uses only category labels during training, thus the amount of data annotations becomes manageable. Among those, Class Activation Mapping (CAM) \cite{zhou2016learning} is a representative weakly-supervised object localization method. CAM is designed to extract a heatmap by analyzing internal layer of CNN, which is then post-processed to locate a bounding box. The main idea of CAM is that the pixels contributing to object classification coincide with the object location. However, in common occasions, even the cheapest information, object categories, may not be affordable.

For a fundamental solution to annotation dependency, unsupervised object localization techniques have emerged. Unlike the weakly-supervised object localization, the unsupervised approach does not rely on any annotations about a dataset. Among the unsupervised approaches, co-localization aims to find a target object in a dataset which contains only one object category. Because the negative samples (e.g., other category objects) are not provided, this problem is inherently more challenging than the weakly-supervised object localization problem.

Unlike supervised, or weakly-supervised methods, unsupervised object co-localization techniques are yet to employ deep neural networks. Existing techniques such as \cite{kim2009unsupervised,rubinstein2013unsupervised,tang2014co,zhu2015unsupervised,cho2015unsupervised} rely on hand-crafted feature extractions, graph-based theories, or optimizations, placing a limit to achieving real-time performance. Meanwhile, deep neural network models are considered outstanding at feature extraction, superior to the previous hand-crafted ones in most pattern recognition problems, even achieving real-time performance. Motivated by their recent success, we aim to apply deep neural networks to unsupervised object co-localization, expecting improvements in both performance and time efficiency. Specifically, in this paper, we suggest an end-to-end unsupervised object co-localization method based on Generative Adversarial Networks (GANs) \cite{goodfellow2014generative} for the first time.

GAN is an unsupervised generative model that learns to generate samples following the true data distribution by implicit density estimation. GAN is composed of a generator and a discriminator. The generator is trained in a way the discriminator cannot distinguish fake images, produced by the generator, from real images. Meanwhile, the discriminator learns to distinguish them from real images. Through this adversarial competition, the generated images from GANs become more realistic, hard to distinguish from the reals. Among many generative models, GAN is known to generate most sharp and realistic images.

In this paper, we take advantage of a GAN discriminator for unsupervised object co-localization. Without utilizing any priors or annotations, GAN successfully generates images that follow the true data distributions. Suppose that the generator is trained to produce a dominant object (i.e., the most frequently appearing object in a dataset), which is our target object in co-localization dataset. Then, we expect that the discriminator will pay more attention to the target object in distinguishing real or fake.

Unfortunately, a GAN discriminator may not always use the target object as a decision criteria. In fact, GANs may generate not only the dominant object, but also other various objects in the dataset. Especially, recent advanced GAN algorithms improve image diversity in data generation. For example, to encourage diverse image generations, Arjovsky et al. \cite{arjovsky2017wasserstein} introduced a novel loss function, and Gulrajani et al. and Kodali et al. \cite{gulrajani2017improved,kodali2017dragan} added a regularization term. In this way, their GAN models capture various modes of data distribution. As a result, their discriminator may learn modes that are not related to the target object. If diverse scene components such as background objects are generated by the GAN generator, its discriminator can properly distinguish those of scene components, thus modeling them as separate modes. This is a desirable property for a GAN model, because achieving the image diversity is an important objective for GAN training. However, for the object localization, this is a prohibitive property since the diversity is negative to capture the target object; the ideal object localization should find the target object only, not other scene components.  

To meet the goal of object localization, we ironically return to the early models of GANs. The early GAN models have been well known to easily miss minor modes of data distribution during training. The consequent phenomenon is called \emph{mode collapse}, a major issue in GAN training.

Although this mode collapse is considered undesirable for GAN based image generation, we expect this pathological behavior is rather useful in our application. When mode collapse occurs, we observed that the most frequently appearing object in the dataset is generated while less frequently appearing objects are lost. Based on this observation, we regard that the target object corresponds to a major mode in a data distribution, while minor modes correspond to less frequently appearing objects, such as background objects. For localization, GAN models need to focus only on the major mode, not the minor modes. Using this as our motivation, we suggest that GAN presenting low diversity in image generation can be utilized for unsupervised object co-localization tasks. We believe our contribution lies at showing such potential for the first time. 

Our model receives a single image as an input, and outputs a heatmap, or a bounding box. The GAN model is trained in an unsupervised manner, and the heatmap is extracted from the discriminator using CAM. Then, the heatmap is post-processed to determine the bounding box for object localization. Within the whole process, neither supervision, nor any extra labeling information, such as negative sampling, is required or used.

By leveraging publicly available datasets, we demonstrate the feasibility of GAN for addressing the problem of unsupervised object co-localization. In addition, we have found that the early GAN model \cite{radford2015unsupervised} is better than state-of-the-arts GAN models \cite{gulrajani2017improved,kodali2017dragan} for localization. Furthermore, we show that quantitative and qualitative performance of our model is even comparable to the ones with weakly-supervised object localization, which utilizes external dataset and corresponding labels. To the best of our knowledge, our proposition is the first end-to-end deep neural network model for unsupervised object co-localization, and we believe that this method can serve as an important baseline for future co-localization researches.

\section{Related works}
\label{sec:relatedworks}

\noindent\textbf{Weakly-supervised object localization.} Weakly-supervised techniques utilize category labels, relatively low-cost annotations, to perform object localization. Traditional weakly-supervised approaches \cite{bilen2014weakly,song2014learning,cinbis2014multi,song2014weakly,li2016weakly,cinbis2017weakly} mostly learn patterns or features that best discriminate object categories from the others, and facilitate such information to localize objects. These methods, however, have a disadvantage that they only capture a fraction of an object, because localization is performed only with the most discriminative part of the object.

Recently, weakly-supervised approaches utilize Deep Convolutional Neural Networks (CNNs) \cite{zhou2016learning,simonyan2013deep,oquab2015object,bilen2016weakly,kantorov2016contextlocnet,selvaraju2017grad}. These approaches analyze the internal layer of CNN classifiers, which are trained with category labels, for object localization. They extract a heatmap, also known as a saliency map, from the CNN classifier, then localize objects by post-processing it. However, similar to the conventional methods, these CNN-based approaches also cover only a fraction, not the whole object. In order to overcome such disadvantage, the latest techniques \cite{kim2017two,singh2017hide,choe2018improved} have removed the most discriminative parts from the image during training so that the network can consider the entire object. However, these weakly-supervised approaches cannot perform object co-localization when negative samples are not available in a dataset, because they rely on the discriminative power to understand the data distribution.

\noindent\textbf{Unsupervised object co-localization.} Unsupervised approaches do not depend on additional annotations for object localization. This problem is very challenging in that there is no other information except the given images. For that, many techniques were designed under an assumption that only one object category exists in a dataset \cite{kim2009unsupervised,rubinstein2013unsupervised,tang2014co,zhu2015unsupervised,cho2015unsupervised,wei2017deep}. Such method is known as co-localization. Recently, Wei et al. \cite{wei2017deep} proposed the usage of the deep features from a CNN, pre-trained using ImageNet dataset \cite{russakovsky2015imagenet}, and calculated their correlation for co-localization. Although this technique utilizes deep neural networks as a feature extractor, this is different from our approach in that we develop an end-to-end network, and do not require a pre-trained network that is trained with an external dataset. For this unsupervised co-localization, our approach adopts GANs \cite{goodfellow2014generative}, the most widely used model among unsupervised deep neural network models, and CAM \cite{zhou2016learning}, one of state-of-the-art techniques for extracting heatmaps of CNN classifiers. We will explain the principle of GAN and CAM as a background of our proposed method in the next section.

\section{Background}
\label{sec:background}

\subsection{Generative Adversarial Networks (GANs)}

Recent approaches to improving GANs aim to achieve either the image quality or the image diversity (i.e., resolving mode collapse). In this subsection, we will review the related work of developing GANs toward image quality and diversity.

Goodfellow et al. originally invented Generative Adversarial Networks (GANs) \cite{goodfellow2014generative}. GAN training proceeds by adversarial competition between a generator $G$ and a discriminator $D$. The discriminator outputs a probability of the input to be real, and is trained by minimizing an objective function shown in Eq.~(\ref{eqn:discobjective}). 

\vspace*{-5mm}
\begin{equation}
    J_{(D)}(\theta_{D},\theta_{G})=
    -\mathbb{E}_{x\sim p_{real}}[\log{D(x)}]
    -\mathbb{E}_{z\sim p_{noise}}[\log{(1-D(G(z)))}]\label{eqn:discobjective}
\end{equation}

The input to the generator, $z$, is sampled from a random noise distribution $p_{noise}$. $x$ denotes real images sampled from dataset. For training GANs, they suggest two different loss functions: a minimax loss, and a non-saturating loss. The minimax loss induces that generated samples are unlikely to be fake by the discriminator, minimizing Eq. (\ref{eqn:minimax}). Meanwhile, the non-saturating loss encourages that generated samples are more likely to be real by discriminator, minimizing Eq.~(\ref{eqn:ns}). 

\begin{equation}
    J_{(G)}(\theta_{G})=
    \mathbb{E}_{z\sim p_{noise}}[\log{(1-D(G(z)))}]\label{eqn:minimax}
\end{equation}

\vspace*{-4mm}
\begin{equation}
    J_{(G)}(\theta_{G})=
    -\mathbb{E}_{z\sim p_{noise}}[\log{(D(G(z)))}]\label{eqn:ns}
\end{equation}

Goodfellow et al. \cite{goodfellow2014generative} used the minimax loss to mathematically prove that their method can achieve Nash equilibrium between the discriminator and generator. However, they suggested that it is better to use non-saturating loss in practical applications due to a gradient vanishing problem of the minimax loss at the early stage of the training. Later, to improve the stability of GAN training, Radford et al. \cite{radford2015unsupervised} found the optimal combination of network architectures using convolution layers and hyperparameters for non-saturating loss. This method is called a Deep Convolutional GAN (DCGAN), the most widely used baseline network architecture in developing GANs.

Later, Arjovsky and Bottou \cite{arjovsky2017towards} have shown that gradient vanishing or mode collapse arose because of the strict condition for convergence of Eqs. (\ref{eqn:minimax}) and (\ref{eqn:ns}). Specifically, the non-saturating objective function can be formulated in the form of the weighted sum of reverse Kullback-Leibler (KL) divergence and Jensen-Shannon (JS) divergence. This reverse KL loss is robust to gradient vanishing, however, easily falls into mode collapse \cite{arjovsky2017towards}. 

To address this problem, Arjovsky et al. \cite{arjovsky2017wasserstein} constructed a Wasserstein distance based loss function, which relaxes the condition for convergence. However, the Wasserstein distance cannot be directly implemented using neural networks without approximation. For approximating the Wasserstein distance, they used Kantorovich-Rubinstein (KR) duality, which requires 1-Lipschitz condition. To this end, they employed a weight clipping scheme for their discriminator to meet the 1-Lipschitz continuity condition. Consequently, Wasserstein GAN (WGAN) successfully stabilize training process and improve image diversity. Eq.~(\ref{eqn:wgan}) shows objective functions of WGAN.

\vspace*{-4mm}
\begin{equation}
\begin{split}
     J_{(D)}(\theta_{D},\theta_{G})=
        \mathbb{E}_{x\sim p_{real}}[D(x)]
        -\mathbb{E}_{z\sim p_{noise}}[D(G(z))], 
        \\ J_{(G)}(\theta_{G})
        =
        -\mathbb{E}_{z\sim p_{noise}}[D(G(z))]\label{eqn:wgan}
        \end{split}
\end{equation}

However, the weight clipping methods cause some problematic behaviors such as image quality degradation. For replacing the weight clipping, Gulrajani et al. \cite{gulrajani2017improved} introduced a Gradient Penalty (GP) term. This method is commonly known as WGAN-GP, and is regarded as one of state-of-the-art techniques for both image diversity and quality. Meanwhile, Kodali et al. \cite{kodali2017dragan} pointed out that GP term proposed by \cite{gulrajani2017improved} may violate the Kantorovich-Rubinstein (KR) duality in certain condition; they proved that the KR duality is valid only if GAN reproduces the true data distribution sufficiently well when GP term proposed by \cite{gulrajani2017improved} is used. For the solution, they suggested a novel GP term that is based on the no-regret algorithm in Game theory. This method is called Deep Regret Analytic GAN (DRAGAN), and is also evaluated as state-of-the-art technique along with WGAN-GP. After that, Fedus et al. \cite{fedus2017many} showed experimentally that both GP terms can improve image diversity of non-saturating loss based GANs, although KR-duality is not neccessary for the non-saturating loss. Both GP terms can be formulated as Eq. (\ref{eqn:gp}). In WGAN-GP, the $x_{gp}$ is equivalent to the generated samples, $G(z)$. In DRAGAN, the $x_{gp}$ are \emph{perturbed} samples, added random Gaussian noise to real samples $x_{real}$. Note that $\alpha$ is a random value sampled from uniform distribution between $[0,1)$.

\vspace*{-2mm}
\begin{equation}
\mathbb{E}_
{\bar{x}\sim p_{\bar{x}}}
[( \parallel \nabla_{\bar{x}} D(\bar{x}) \parallel_2 -1)^2]
,
\quad
\bar{x}=\alpha x_{real}+(1-\alpha)x_{gp} \label{eqn:gp}
\end{equation}

Recently, Miyato et al. \cite{miyato2018spectral} introduced a novel weight normalization method called a spectral normalization (SN), so that the discriminator can meet the 1-Lipchitz continuity. SN normalizes the weights of all trainable layers respectively in discriminator by their spectral norms which is the largest singular value. The SN stabilize the GAN training process and improve the quality of generated samples of GANs. Similar to gradient penalty, SN improves not only Wasserstein distance based GANs, but also non-saturating loss based GANs \cite{miyato2018spectral}. It is interesting because the 1-Lipschitz continuity is only required for KR-duality which is utilized by WGAN, not required for non-saturating loss based GANs.

\subsection{Class Activation Mapping (CAM)}

CAM \cite{zhou2016learning} visualizes the region where the CNN classifier is influenced to decide an object category. The motivation of CAM comes from an insight that discriminative features are closely related to the object's appearance frequency, and its location. For instance, CNN classifiers learn discriminative features, characterized by the appearance frequency of the target objects; a target object always exists for corresponding class images, whilst unlikely to appear for other class images. These discriminative features generally reside inside the object, where CAM trace them for object localization.

Specifically, CAM replaces fully-connected layers of the CNN with Global Average Pooling (GAP) layer. In this way, the spatial information of the image is preserved to the last softmax layer. The weight associating the last GAP layer and the classification layer indicates how important the each activation map of the last convolutional layer just before the GAP layer is. Exploiting this weight information, a heatmap of the target category can be extracted by the weighted sum of activation maps from the last convolutional layers. This heatmap is binarized by simple thresholding with pre-determined ratio of the maximum value. After that, they exam the connected components from binarized heatmap, draw the tightest bounding box for each connected component, and select the largest box.

\section{Co-localization using Generative Adversarial Networks}
\label{sec:methods}
GAN discriminator extracts features to learn the best distinction between real and fake. Especially, the discriminative features should encapsulate the information that distinguishes the target object from the others. Since discriminative features of CNN classifiers can be utilized for object localization as CAM \cite{zhou2016learning} pointed out, we also argue that object location can be extracted using discriminative features from GAN discriminator, a CNN binary classifier. 

\noindent\textbf{Proposed approaches.} We add a Global Average Pooling (GAP) layer at the end of the last convolutional layer of the GAN discriminator, and connect this GAP layer to the binary classification layer in a fully-connected manner. Note that the weight between the GAP layer and the classification layer of the CAM indicates how much each activation map of the last convolution layer contributes to decide the category label. Likewise, our weight represents how much each activation map contributes in correctly distinguishing real and fake images. The extracted heatmap from such trained discriminator can result in the bounding box by following the same post-processing technique applied by CAM. We also assume that there is only a single object in a image like other weakly-supervised approaches \cite{zhou2016learning,kim2017two,singh2017hide,choe2018improved}. This means the proposed approach will draw only one bounding box per image.

\noindent\textbf{Selected GANs.} Recent GAN models have made a meaningful progress toward improving image diversity. In other words, they successfully encapsulate most modes of data distribution \cite{arjovsky2017wasserstein,gulrajani2017improved,kodali2017dragan,fedus2017many}. In such case, the discriminator is likely to consider the overall region of image for discriminating. It is because it learns not only the dominant object, but also other components such as background objects, which appears relatively less in the dataset. Those GANs are superior to the early models of GAN in the perspective of image diversity.

On the other hand, early models of GAN are known to easily fall into mode collapse. These models easily miss less frequently appearing objects which corresponds to minor modes, and only generate the dominant object in the dataset. We argue that it is because the discriminator does not utilize any other information except for that of target object as a decision criteria. Although this phenomenon might be disadvantageous for image diversity, this is a desirable property for object localization. Therefore we believe that low diversity can be positive for object localization.

To verify our argument, we select three variants of GAN models: DCGAN \cite{radford2015unsupervised}, WGAN-GP \cite{gulrajani2017improved}, and DRAGAN \cite{kodali2017dragan}, with application of spectral normalization \cite{miyato2018spectral} to stabilize GAN training. Specifically, DCGAN is known to easily fall into mode collapse, while WGAN-GP \cite{arjovsky2017wasserstein} and DRAGAN \cite{kodali2017dragan} are more robust to mode collapse, as we mentioned in Sec. \ref{sec:background}. Therefore, in order to investigate how the image diversity of GAN affects object localization, we compare DCGAN with WGAN-GP and DRAGAN.

\noindent\textbf{Upper-limit reference.} We note that the CAM is not applicable for the co-localization problem; co-localization dataset does not provide negative samples. Because CNNs cannot be trained only with positive samples, the CAM utilizes external negative dataset including their labels as additional information for training. Thus, it is obvious that CAM should perform better than the unsupervised co-localization. Therefore, we argue that CAM should serve as our upper-limit reference. 

\section{Implementation Details}
\label{sec:details}

\noindent\textbf{Network architecture.} We decide hyperparameters and network architectures based on DCGAN \cite{radford2015unsupervised}. In addition, for extracting heatmap, we add a GAP layer right after the last convolution layer of discriminator. Table \ref{tab:architecture} shows the details of our standard networks architecture for 64$\times$64 inputs. The input of generator, $z$, is sampled from uniform distribution between $[-1,1)$. We use $\alpha = 0.2$ for leaky ReLU activation.

We modify the networks in Table \ref{tab:architecture} according to loss functions, input size, and usage of spectral normalization. For 32$\times$32 images, we set the stride to 1 of first convolution layer of discriminator and last transpose convolution layer (denoted as Deconv in Table \ref{tab:architecture}) of generator. The spectral normalization \cite{miyato2018spectral} is only applied to the discriminator. For SN-DCGAN, we remove batch normalization \cite{ioffe2015batch} and apply spectral normalization to all convolution and fully-connected layers of DCGAN. On the other hand, we do not apply spectral normalization to DRAGAN, because DRAGAN with SN cannot generate any meaningful image. For WGAN-GP, we use layer normalization \cite{ba2016layer} instead of batch normalization according to the original paper. To implement SN-WGAN-GP, we apply spectral normalization to all layers of WGAN-GP. Note that we keep layer normalization for SN-WGAN-GP together with spectral normalization, because it cannot be trained without layer normalization.

\begin{table}[t!]
\centering
\caption{Our standard network architectures for 64$\times$64 inputs. We modify this network architecture according to objective functions, input size, and usage of spectral normalization.}
\label{tab:architecture}
\begin{scriptsize}
\begin{tabular}{@{}llll@{}}
\toprule
\multicolumn{2}{c}{Generator}                                           & \multicolumn{2}{c}{Discriminator}                                      \\ \midrule
\multicolumn{1}{c}{Layers}           & \multicolumn{1}{c}{Output shape} & \multicolumn{1}{c}{Layers}          & \multicolumn{1}{c}{Output shape} \\ \midrule
z                                    & 128                              & Conv, 4$\times$4, 2, 64, BN, lReLU  & 32$\times$32$\times$64           \\ \midrule
Linear                               & 4$\times$4$\times$128            & Conv, 4$\times$4, 2, 128, BN, lReLU & 16$\times$16$\times$128          \\ \midrule
Deconv, 4$\times$4, 2, 512, BN, ReLU & 8$\times$8$\times$512            & Conv, 4$\times$4, 2, 256, BN, lReLU & 8$\times$8$\times$256            \\ \midrule
Deconv, 4$\times$4, 2, 256, BN, ReLU & 16$\times$16$\times$256          & Conv, 4$\times$4, 2, 512, BN, lReLU & 4$\times$4$\times$512            \\ \midrule
Deconv, 4$\times$4, 2, 128, BN, ReLU & 32$\times$32$\times$128          & Global Average Pooling              & 1$\times$1$\times$512            \\ \midrule
Deconv, 4$\times$4, 2, 3, Tanh       & 64$\times$64$\times$3            & Linear                              & 1                                \\ \bottomrule
\end{tabular}
\end{scriptsize}
\end{table}

\noindent\textbf{Training process.} Eqs. (\ref{eqn:discobjective}) and (\ref{eqn:ns}) are iteratively minimized for training DCGAN. For DRAGAN, while the original paper uses Eqs. (\ref{eqn:discobjective}), (\ref{eqn:minimax}) and (\ref{eqn:gp}) for training according to the original paper \cite{kodali2017dragan}, we use Eq. (\ref{eqn:ns}) instead of Eq. (\ref{eqn:minimax}) as \cite{fedus2017many} recommended. For WGAN-GP, Eqs. (\ref{eqn:wgan}) and (\ref{eqn:gp}) are minimized for training. We update the discriminator once per one update of generator for DCGAN and DRAGAN, and update the discriminator five times per one update of generator for WGAN-GP. The batch size of all our experiments is 128 and the training iteration is 250k. We use an Adam optimizer, with $\beta_{1}$ of 0.9 and $\beta_{2}$ of 0.999. The learning rate is 2e-4.

\noindent\textbf{Data augmentation.} Data augmentation increases the amount of training data by perturbing the data. Specifically, we apply small translation randomly to input images by 5\% of their size, and modify brightness, contrast, saturation, and lighting \cite{krizhevsky2012imagenet}. This data augmentation is known to stabilize the network training process. Our experimental study shows how these techniques influence the GAN training and the localization performance.

\noindent\textbf{Class Activation Mappings.} Although the original implementation of CAM is built upon AlexNet \cite{krizhevsky2012imagenet} and GoogLeNet \cite{szegedy2015going}, we replace the baseline CNNs with pre-activation ResNet \cite{he2016identity} to improve the CAM. Pre-activation ResNet is one of state-of-the-art CNN classification networks. We choose the 34-layer architecture with batch size of 256 and fix the training iteration with 100k. We also follow the original paper \cite{he2016identity} to decide hyperparameters for implementing the ResNet-34. Specifically, we use a momentum optimizer, with momentum of 0.9. Learning rate is initially set to 0.1, decayed by a factor of 10 every 25k iterations. The weight decay is 1e-4. 

\section{Experimental Results}
\label{sec:experimentalresults}
\begin{table}[t!]
\centering
\caption{The list of dataset for co-localization experiments.}
\label{tab:codataset}
\begin{tabular}{lccl}
\Xhline{3\arrayrulewidth}
\multicolumn{1}{c}{Name} & \multicolumn{1}{l}{\# train} & \# test & \multicolumn{1}{c}{ImageNet subcategory name}                                                                                                       \\ \hline\hline
Artiodactyla             & 2500                         & 250     & bison, ox, bighorn, gazelle, arabian camel                                                                                                 \\ 
Bottle                   & 1000                         & 100     & pop bottle, beer bottle                                                                                                                    \\ 
Bird                     & 1500                         & 150     & albatross, black stork, goose                                                                                                              \\ 
Cat                      & 2000                         & 200     & tabby, persian cat, egyptian cat, cougar                                                                                                   \\ 
Dog                      & 3000                         & 300     & \begin{tabular}[c]{@{}l@{}}standard poodle, yorkshire, golden retriever,\\ labrador retriever, german shephered, chihuahua\end{tabular}     \\ 
Vehicle                  & 4000                         & 400     & \begin{tabular}[c]{@{}l@{}}convertible, school bus, trolleybus, sports car, \\ police van, moving van, limousine, beach wagon\end{tabular} \\ \Xhline{3\arrayrulewidth}
\end{tabular}
\end{table}

\begin{figure}[t!]
    \noindent
    \begin{center}
        \subfigure[SN-DCGAN]{\includegraphics[width=0.32\columnwidth]{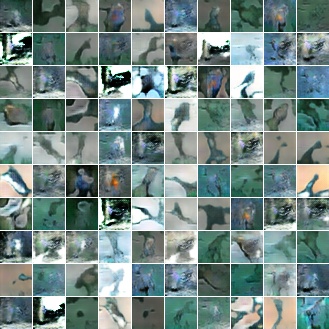}}
        \subfigure[DRAGAN]{\includegraphics[width=0.32\columnwidth]{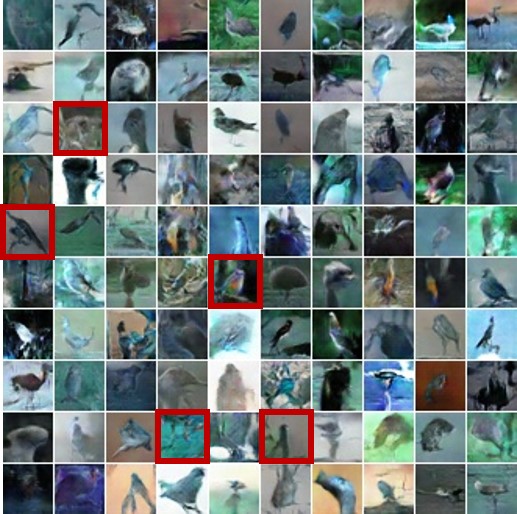}}
        \subfigure[SN-WGAN-GP]{\includegraphics[width=0.32\columnwidth]{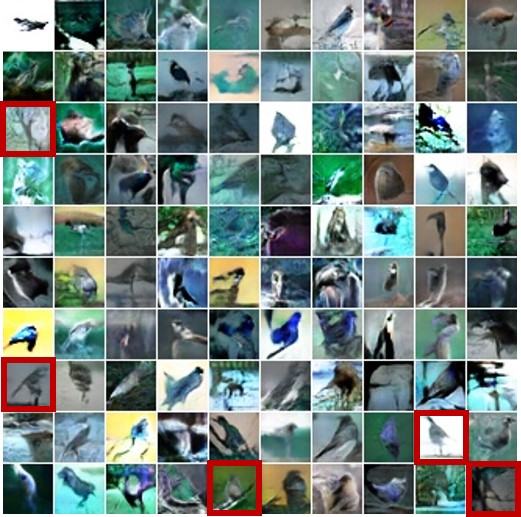}}
    \end{center}
    \vspace*{-5mm}
    \caption{Generated samples of each GAN model on the \emph{bird} class of CIFAR-10 at same iteration. SN-DCGAN samples successfully simplify the object appearance while the samples from DRAGAN and SN-WGAN-GP show the complex appearance. We emphasis generated background objects using red boxes.  Although the limited diversity of DCGAN is interpreted as the weakness in developing GANs, it is beneficial to our applications.}
\label{fig:cifar10gen}
\vspace*{-5mm}
\end{figure}

\noindent\textbf{Dataset.} We use low-resolution datasets, such as CIFAR-10 (32$\times$32$\times$3) and Tiny ImageNet (64$\times$64$\times$3), a reduced version of ImageNet. CIFAR-10 includes 10 categories, and with 5000 training images and 1000 test images per category. We perform object co-localization on each category of CIFAR-10. In Tiny ImageNet, there are 200 categories in total, and each category consists 500 training images and 50 test images. We combine similar categories in Tiny ImageNet and rearrange them for constructing six co-localization datasets. Table \ref{tab:codataset} summarizes the list of each co-localization dataset using Tiny ImageNet with the total number of images.

Although ImageNet or MS-COCO dataset is the most popular benchmark dataset for recent object recognition and localization techniques, handling such high resolution general object datasets with GAN is still an open problem. Therefore, we leave it for a future work.

\noindent\textbf{Evaluation metric.} We use the quantitative evaluation metrics as suggested in a previous work \cite{singh2017hide}. Specifically, we utilize the localization accuracy with a known ground-truth class (\textit{GT-known Loc}). This metric marks correct, only when the intersection of union (IoU) between estimated bounding box and the ground truth box is greater than 50\%. We also perform qualitative evaluation by visualizing the bounding box and the heatmap. In the case of CIFAR-10 experiments, we only conduct qualitative evaluation, because ground truth bounding box annotations are not available. For Tiny ImageNet experiments, we analyze the results both qualitatively and quantitatively.

\begin{figure}[t!]
    \noindent
    \begin{center}
        \subfigure[SN-DCGAN (Proposed)]{\includegraphics[width=1\columnwidth]{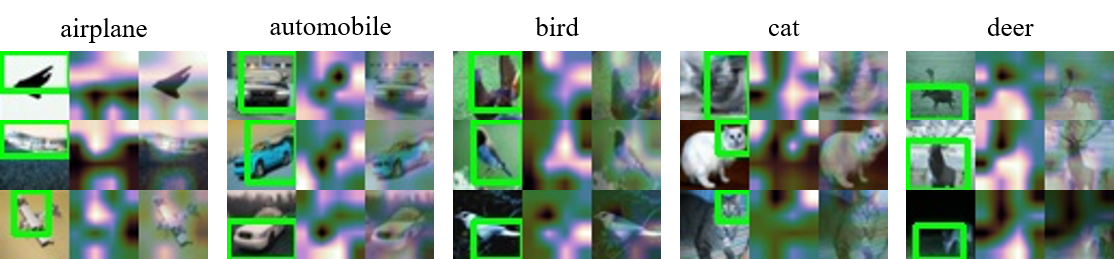}}\\
        \subfigure[DRAGAN]{\includegraphics[width=1\columnwidth]{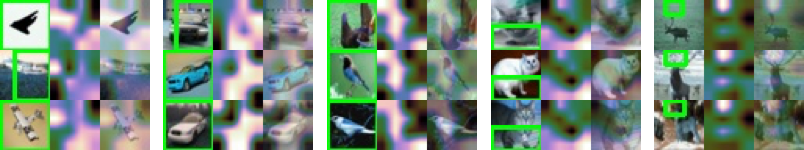}}\\
        \subfigure[SN-WGAN-GP]{\includegraphics[width=1\columnwidth]{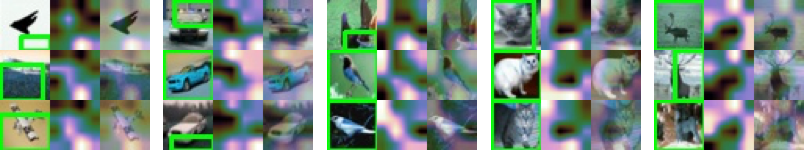}}\\
    \end{center}
    \vspace*{-5mm}
    \caption{Object localization results on CIFAR-10. In each subfigure, the left column is the input image with green estimated bounding box, the middle is the extracted heatmap, while the right column is the result of overlapping heatmap and an input. We observe that heatmaps from SN-DCGAN successfully localize the main objects. On the other hand, for every class, the heatmaps of SN-WGAN-GP and DRAGAN produce nearly identical results regardless of the input image. This suggests that SN-DCGAN may not be able to generate various objects of dataset, but it can be advantageous for object localization. Note taht we did not visualize the results of all categories due to limited space.}
\label{fig:cifar10}
\vspace*{-5mm}
\end{figure}

\begin{figure}[h!]
    \noindent
    \begin{center}
        \subfigure[Four-legs animals]{\includegraphics[width=0.9\columnwidth]{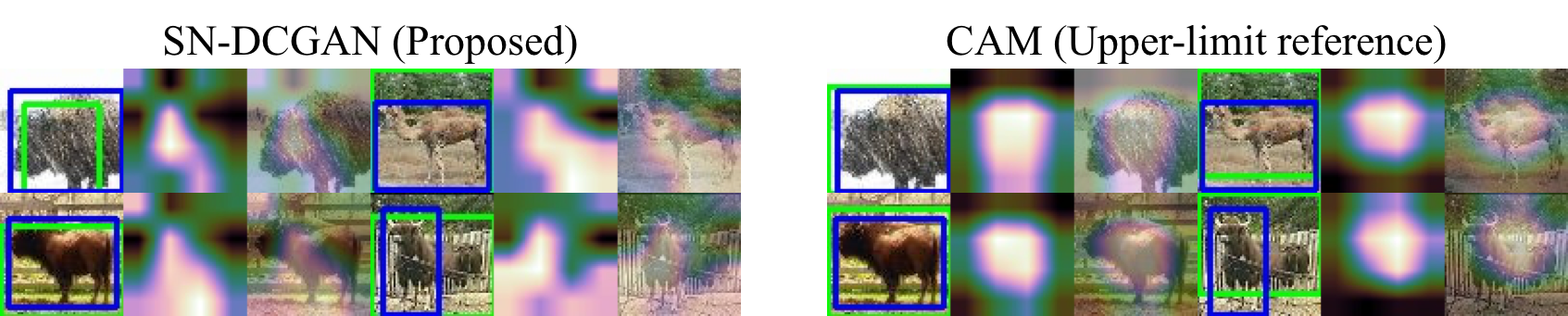}}\vspace*{-1.5mm}
        \subfigure[Bird]{\includegraphics[width=0.9\columnwidth]{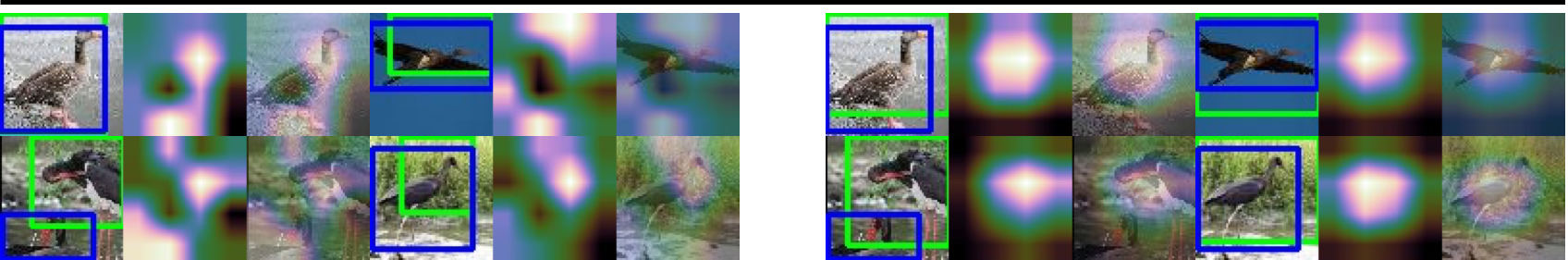}}\vspace*{-1.5mm}
        \subfigure[Bottle]{\includegraphics[width=0.9\columnwidth]{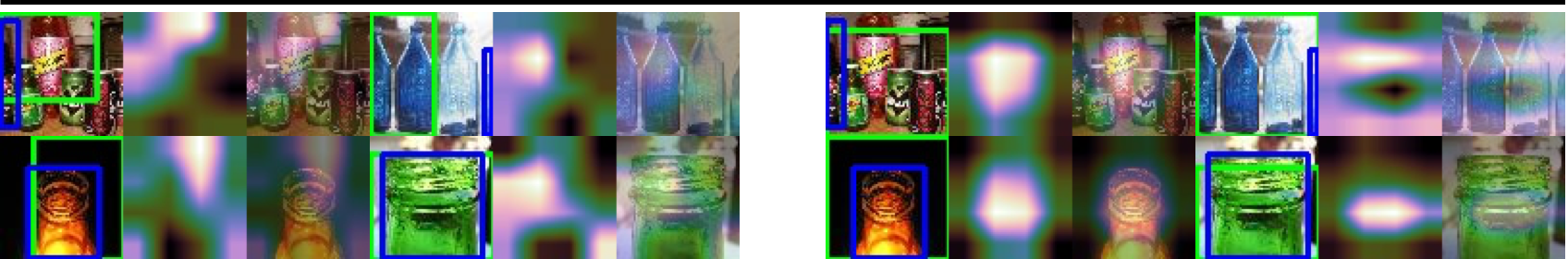}}\vspace*{-1.5mm}
        \subfigure[Cat]{\includegraphics[width=0.9\columnwidth]{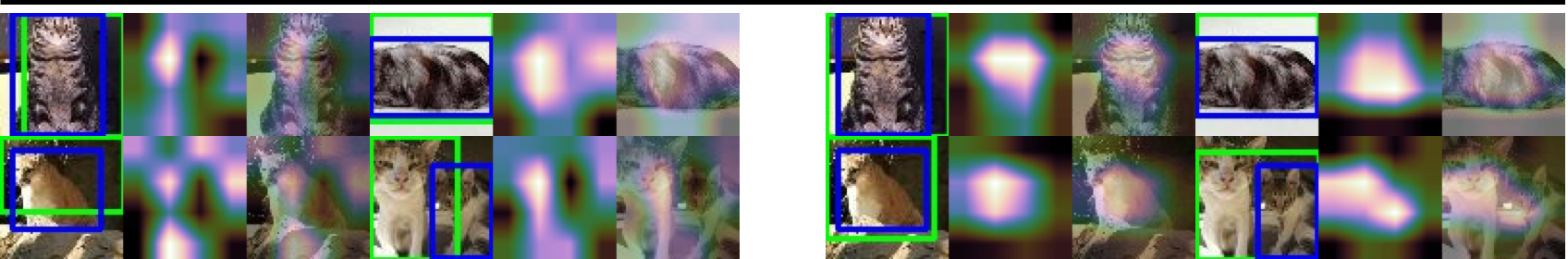}}\vspace*{-1.5mm}
        \subfigure[Dog]{\includegraphics[width=0.9\columnwidth]{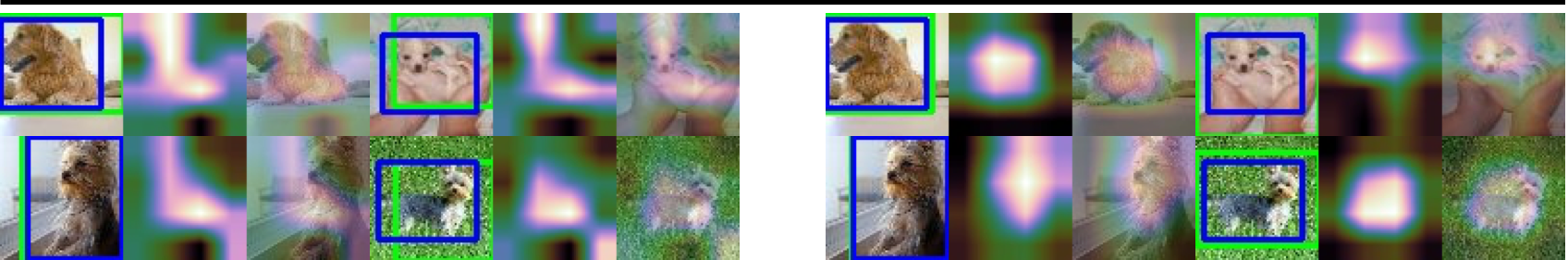}}\vspace*{-1.5mm}
        \subfigure[Vehicle]{\includegraphics[width=0.9\columnwidth]{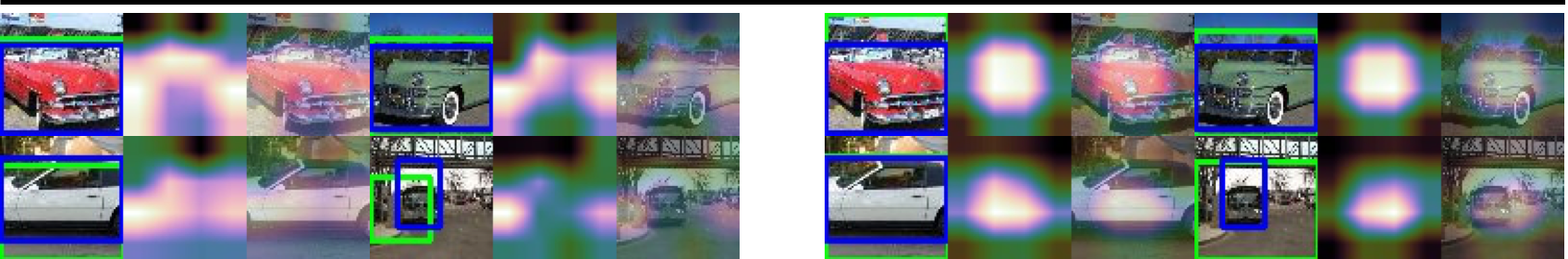}}
    \end{center}
    \vspace*{-5mm}
    \caption{Experimental results on Tiny ImageNet. The blue bounding box is ground truth while green bounding box is our estimates. Note that the localization performance of DCGAN is comparable to SN-DCGAN, while DRAGAN, WGAN-GP, and SN-WGAN-GP have a same problem as CIFAR-10 experiments. However, we do not show their results in this figure due to limited space.}
\label{fig:colocalization}
\vspace*{-5mm}
\end{figure}

\noindent\textbf{CIFAR-10.} We perform object localization using DCGAN, SN-DCGAN, DRAGAN, WGAN-GP, and SNWGAN-GP on each category of CIFAR-10 dataset. We visualizes the generated samples of each GAN on bird category in Fig. \ref{fig:cifar10gen}. In this experiment, we observe that SN-DCGAN focuses on generating only birds by ignoring various background, such as stones or trees, where the birds sit on. Meanwhile, DRAGAN and SN-WGAN-GP even produces various background textures, stones, or trees. Reproducing those of diverse details in DRAGAN and SN-WGAN-GP is a desirable property in GAN training because the image diversity is one of important goal for generative models. For object co-localization, however, the information other than the target object should be neglected because they are outliers in localization problem. Note that we have found that DCGAN and WGAN-GP have same phenomenon shown by their SN version respectively, but did not visualize in this paper due to the limited space. 

Fig. \ref{fig:cifar10} shows object localization results using each GAN model; In each subfigure, the first column shows the input with the green estimated bounding box, the second column illustrates heatmaps, and the third presents overlaps of the input with heatmaps. The localization results from SN-DCGAN show that the bounding boxes reasonably capture the locations of the dominant objects, also consistent with heatmaps, which significantly overlaps with the objects. However, for each class, the heatmaps from the DRAGAN and SN-WGAN-GP appear almost identical regardless of input images.

We believe that this phenomenon is caused by characteristics of the objective function for each GAN, and the effect of the GP term. The Wasserstein distance used in WGAN-GP is known to improve the diverse image generation \cite{arjovsky2017wasserstein}. In addition, the GP terms used in WGAN-GP and DRAGAN leads GAN to learn various modes of the data distribution \cite{fedus2017many}. Therefore, we believe that this is why the heatmaps from WGAN-GP and DRAGAN do not vary much upon the input images; the discriminator is developed to exam all objects, thus heatmaps do not focus on the dominant object. Based on these results, for localization, we argue that DCGAN and SN-DCGAN are more appropriate than DRAGAN, WGAN-GP, and SN-WGAN-GP.

Because CIFAR-10 dataset does not have ground truth of object location, we only perform qualitative evaluation in these experiments. For more rigorous verification, we perform quantitative evaluation using Tiny ImageNet dataset.

\begin{table}[t!]
\centering
\caption{Quantitative evaluation results for co-localization experiments using GT-known Loc (\%). The plane numbers mean that the GAN shows the low diversity in image generation while the under-bar numbers indicate that the GAN produces the diverse objects other than the target object. In addition, we use bold text for emphasizing our best score and CAM score.}
\label{tab:colocalization}
\begin{tabular}{@{}lccccc@{}}
\toprule

\multicolumn{1}{c}{\multirow{2}{*}{Dataset}} & \multicolumn{2}{c}{DCGAN} & \multicolumn{2}{c}{SN-DCGAN} & CAM    \\ \cmidrule(l){2-6} 

\multicolumn{1}{c}{}                         & \scriptsize w/o Augment.           & \scriptsize w/ Augment.    & \scriptsize w/o Augment.  & \scriptsize w/ Augment.       & \scriptsize w/ Augment. \\

\midrule
Four-legs animals                            & 41.4         & 43.0                & \textbf{54.4}              & \underbar{44.0}       & \textbf{53.4}   \\
Bird                                         & 49.4         & 52.3                & 58.0                       & \textbf{60.6}         & \textbf{57.0}   \\
Bottle                                       & 36.0         & 36.4       & 38.0                       & \textbf{41.0}                  & \textbf{34.3}   \\
Cat                                          & 66.3         & 71.8                & \textbf{78.0}              & 76.5                  & \textbf{80.4}   \\
Dog                                          & 55.0         & 60.8                & \textbf{63.0}              & \textbf{63.0}         & \textbf{65.5}   \\
Vehicle                                      & 54.0         & 61.7                & \textbf{75.0}              & 68.0                  & \textbf{74.1}   \\\bottomrule
\end{tabular}
\end{table}

\begin{figure}[t!]
    \noindent
    \begin{center}
        \subfigure[w/o augmentation]{\includegraphics[width=0.45\columnwidth]{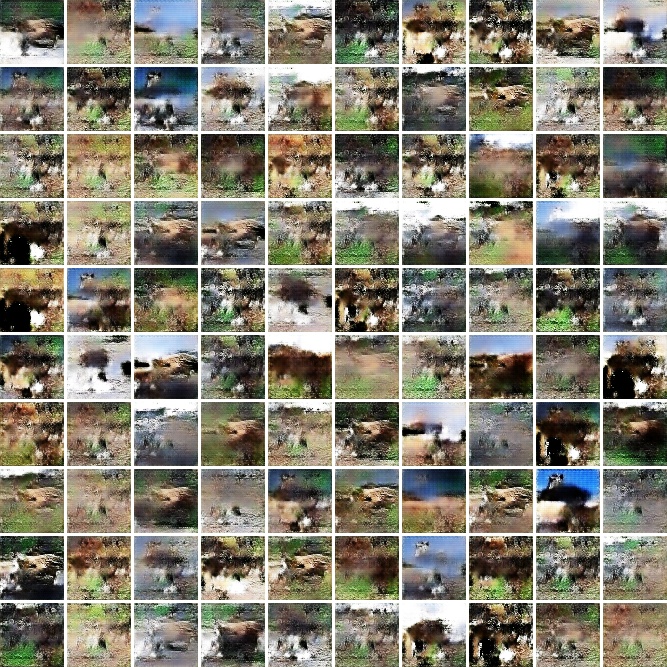}}\quad
        \subfigure[w/ augmentation]{\includegraphics[width=0.45\columnwidth]{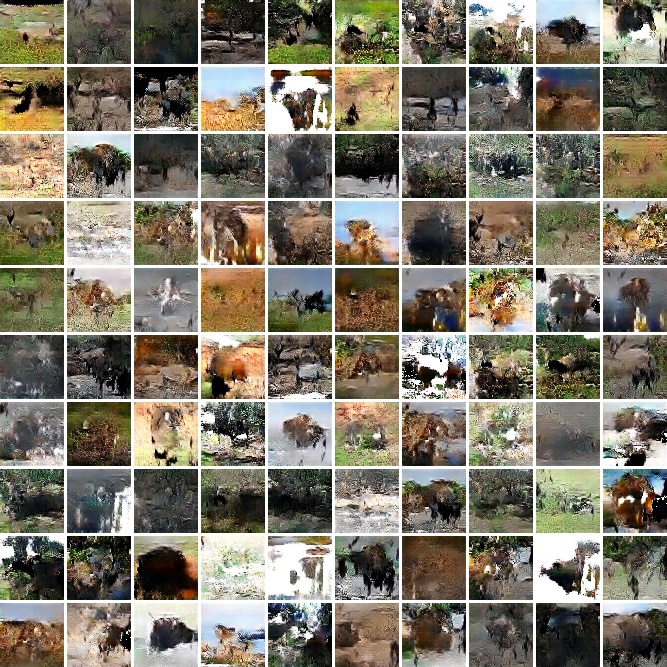}}
    \end{center}
    \vspace*{-5mm}
    \caption{Generated samples of SN-DCGAN on \emph{Four-legs animals} category. We can observe the image diversity of SN-DCGAN with augmentation is higher than that of SN-DCGAN without augmentation. Although the image quality is not good, we can observe that left image shows only main target object while right image shows various objects other than target object.}
\label{fig:augmentation}
\vspace*{-5mm}
\end{figure}

\noindent\textbf{Tiny ImageNet.} We conduct object localization with six co-localization datasets using DCGAN, SN-DCGAN, DRAGAN, WGAN-GP, and SN-WGAN-GP, respectively. Among five GAN models, we only demonstrate the results from SN-DCGAN in Fig. \ref{fig:colocalization} because it is our best model. In fact, as we mentioned in CIFAR-10 experiments, heatmaps from DRAGAN, WGAN-GP, and SN-WGAN-GP yield similar shape regardless of the input, thus they are not meaningful for localization problem. In these experiments, we observe that both DCGAN and SN-DCGAN discriminators pay attentions to the location of a dominant object in the image as we intended. We also provide the results from weakly-supervised approach (CAM \cite{zhou2016learning}). Again, it is important to note that CAM is not our competitor, instead serves as a upper limit as discussed in Sec.~\ref{sec:methods}.

Table \ref{tab:colocalization} shows the quantitative evaluation results of DCGAN, SN-DCGAN, and CAM. Impressively, our co-localization results show comparable performance compared to the weakly-supervised approach. Occasionally, our approach shows even superior performance in the Four-legs animals, Bird, Bottle, and Vehicle categories. We consider such results meaningful, given that the CAM utilizes additional data and corresponding category labels for localization. In addition, we have observed that the spectral normalization improves the localization performance consistently. We believe that this is because the stabilized GAN training due to the spectral normalization is advantageous for localization.

The under-bar text in the table indicates the case where we observe that the GAN produced various objects other than the target object. On the other hand, the plane text corresponds to the case where we observe that the GAN only produced the target object. We visualize the generated samples of Four-legs animals in Fig. \ref{fig:augmentation}. From the results of Four-legs animals category, we found that the usage of data augmentation may affect the image diversity. Clearly, we can observe that the localization performance has dropped with high image diversity. Based on these results, we confirmed that low diversity is desirable for object localization using GANs.

\section{Conclusions}
\label{sec:conclusions}

In this paper, we propose a novel end-to-end approach for unsupervised object co-localization. To this end, we utilize the GAN discriminator for localizing the target object in an unsupervised manner. Our most important finding is that there is a negative correlation between the image diversity and object localization when adopting the GAN model into the object localization problem. Furthermore, various experimental studies show that our approach has achieved meaningful performance for object localization in both qualitative and quantitative evaluations. 

\noindent\textbf{Limitation.} Currently, we do not provide a stop criteria for training GANs in our application. Although the popular metric such as inception, FID or MS-SSIM are used for measuring the quality and diversity of image generation, those values vary significantly upon the dataset. As a result, those metrics are not reliable as a stop criteria for GAN training. Likewise, our approach has the same weakpoint. Although we observe that MS-SSIM \cite{odena2016conditional} and localization performance are positively related with each other, it is too sensitive due to its random sampling and changes greatly upon the dataset, so it is not appropriate for stop criteria. Therefore, we report the peak accuracy for evaluating the localization accuaracy in this paper. However, upon the progress to develop the better metric for measuring the image diversity, we expect this problem to be solved in the future.

\noindent\textbf{Future work.} The objective of our study may be opposite to regular GAN models, because low image diversity is desirable for GAN-based object localization. Therefore, we plan to design a novel GAN specialized for object localization. Furthermore, we will extend our framework to fully-unsupervised object localization by adopting auxiliary networks that can filter minor objects and preserve only target objects.

\bibliographystyle{splncs}

\end{document}